\documentclass[a4paper,twoside]{article}

\usepackage{epsfig}
\usepackage{subcaption}
\usepackage{calc}
\usepackage{amssymb}
\usepackage{amstext}
\usepackage{amsmath}
\usepackage{amsthm}
\usepackage{multicol}
\usepackage{pslatex}
\usepackage{multirow}
\usepackage{algorithm2e}
\usepackage[table]{xcolor}
\usepackage[bottom]{footmisc}
\usepackage[style=apa]{biblatex}
\usepackage{listings}
\usepackage{hyperref}
\usepackage{float}
\usepackage{SCITEPRESS}     

\lstdefinestyle{mystyle}{  
    numberstyle=\footnotesize\bfseries,
    basicstyle=\ttfamily\footnotesize,   
    breaklines=true,                 
    captionpos=b,              
    numbers=left,                  
    numbersep=3pt,
    tabsize=1,
    breakindent=10pt,
    xleftmargin=0.25cm
}

\begin{document}

\title{Enhancing Summarization Performance through Transformer-Based  Prompt Engineering in Automated Medical Reporting}

\author{\authorname{Daphne van Zandvoort\sup{1}\orcidAuthor{0009-0003-7143-6696}, Laura Wiersema\sup{1}\orcidAuthor{0009-0007-8621-3611}, Tom Huibers\sup{2}, Sandra van Dulmen\sup{3}\orcidAuthor{0000-0002-1651-7544}, \\
and Sjaak Brinkkemper\sup{1}\orcidAuthor{0000-0002-2977-8911}}
\affiliation{\sup{1}Department of Information and Computing Sciences, Utrecht University, Utrecht, the Netherlands}
\affiliation{\sup{2}Verticai, Utrecht, the Netherlands}
\affiliation{\sup{3}Nivel (Netherlands institute for health services research), Utrecht, Netherlands}
\email{\{D.W.H.vanZandvoort, L.J.Wiersema, S.Brinkkemper\}@uu.nl, TomHuibers@verticai.nl, S.vanDulmen@nivel.nl}
}


\abstract{Customized medical prompts enable Large Language Models (LLM) to effectively address medical dialogue summarization. The process of medical reporting is often time-consuming for healthcare professionals. Implementing medical dialogue summarization techniques presents a viable solution to alleviate this time constraint by generating automated medical reports. The effectiveness of LLMs in this process is significantly influenced by the formulation of the prompt, which plays a crucial role in determining the quality and relevance of the generated reports. In this research, we used a combination of two distinct prompting strategies, known as \textit{shot prompting} and \textit{pattern prompting} to enhance the performance of automated medical reporting. The evaluation of the automated medical reports is carried out using the ROUGE score and a human evaluation with the help of an expert panel. The two-shot prompting approach in combination with scope and domain context outperforms other methods and achieves the highest score when compared to the human reference set by a general practitioner. However, the automated reports are approximately twice as long as the human references, due to the addition of both redundant and relevant statements that are added to the report. 
}

\onecolumn \maketitle \normalsize \setcounter{footnote}{0} \vfill

\section{\uppercase{Introduction}}
\label{sec:introduction}

The application of Artificial Intelligence (AI), notably Machine Learning (ML), to enhance healthcare and assist medical decision-making is a rapidly growing field \parencite{hicks2022evaluation}. Large Language Models (LLM) are effectively tackling challenging healthcare tasks, such as disease diagnosis, treatment planning, and medical reporting, using personalized medical prompts, even with limited data \parencite{wang2023prompt}. Prompt engineering in the medical domain, including classification, data generation, anomaly detection, content augmentation, question answering, and medical inference, is crucial in improving these healthcare outcomes \parencite{wang2023prompt}. Ensuring high levels of accuracy and reliability in these AI-driven healthcare applications is essential for their successful integration into medical support systems \parencite{balagurunathan2021requirements}.

Expanding on the role of AI and ML in healthcare, Electronic Health Records (EHRs) have become a pivotal focus, revolutionizing medical data management and communication \parencite{coorevits2013electronic}. EHR documentation has led to significant changes in medical practice with an increase in data access and communication among medical professionals compared to paper records \parencite{overhage2020physician}. However, one significant challenge has been the time-consuming data input and hindrances to in-person patient care, resulting in professional dissatisfaction \parencite{friedberg2014factors}. In response, to lessen this administrative burden, automation of this process was developed by several research initiatives, as demonstrated with the Systematic Literature Review of \cite{van2021digital}. Care2Report (C2R) is the only scientific initiative that focuses on the Dutch medical field and automates medical reporting by utilizing multimodal consultation recordings (audio, video, and Bluetooth), enabling knowledge representation, ontological dialogue interpretation, report production, and seamless integration with electronic medical record systems \parencite{maas2020care2report, elassy2022semi}. This automated medical reporting serves as a prime example of prompt engineering, specifically in the domain of medical dialogue summarization (MDS), illustrating how technology can streamline healthcare processes. 

In automated MDS, the generation of automated medical reporting relies on utilizing state-of-the-art LLMs like Generative Pre-trained Transformers (GPT). The level of detail and specificity in the prompts directly influences the model's comprehension and its ability to produce the expected results \parencite{Bigelow_2023,heston2023prompt,Robinson_2023}. 
Several articles on the effective crafting of prompts, emphasize the significance of context and clarity in the prompts, including the provision of additional relevant information for optimal results \parencite{Bigelow_2023,Robinson_2023}.

Although prompt engineering has a substantial impact on the performance of LLMs, its full potential in the domain of medical problem-solving remains largely unexplored. Thus, this research aims to answer the following research question:\\

\textbf{RQ:} \textit{Which prompt formulation detail yields high performance in automated medical reporting?}\\

To answer the research question we focus on prompt engineering related to automated medical reporting. First, we reviewed existing literature for research within prompt engineering, automatic text summarization, and medical dialogue summarization (Section \ref{sec:relatedwork}). Subsequently, Section \ref{sec:studydesign} reports on prompt formulation, execution, and analysis. The findings are presented and discussed (Section \ref{sec:findingsanddiscussion}). Finally, the work is summarized and suggestions are provided for future work (Section \ref{sec:conclusion}).

\section{\uppercase{Related Work}}
\label{sec:relatedwork}
This study builds on prior research in the realm of prompt engineering, aiming to employ diverse prompting methodologies for generating automated medical reports within MDS, a subset of Automatic Text Summarization (ATS).

\subsection{Prompt Engineering}
A human-initiated prompt serves as the initial step for GPT in comprehending the context and meeting user expectations by producing the desired output \parencite{white2023prompt}. 
This process includes designing, implementing, and refining prompts to optimize their efficacy in eliciting this intended result \parencite{heston2023prompt}. An example prompt in the context of this work is shown in Listing \ref{lst:prompt}.

\begin{lstlisting}[caption=Example of a prompt for automated medical reporting. The example is based on existing research of the C2R program., label=lst:prompt]
You are a bot that generates a medical report in SOAP format based on a conversation between a doctor and a patient. 
Only extract information from the conversation to produce the Subjective, Objective, Analysis, and Plan sections of the medical report
\end{lstlisting}

Based on literature, we decided to use the shot prompting and pattern prompting methods to achieve the highest-performing output since these provide an opportunity to demonstrate an example of the expected output and to delineate the context.

\subsubsection{Shot Prompting}
In-context learning is a method where language models learn tasks through a few examples provided as demonstrations \parencite{dong2022survey}. \textit{Shot prompting} employs in-context learning to guide the model's output. There are three strategies: zero-shot, one-shot, and few-shot prompting \parencite{Anil_2023}. \textit{Zero-shot prompting}, also known as direct prompting, involves giving the model a task without specific examples, relying solely on the knowledge acquired during training \parencite{Anil_2023}. In contrast, \textit{one-shot} and \textit{few-shot prompting} provide examples or `shots' to the model at run-time, serving as references for the expected response's structure or context \parencite{Anil_2023, reynolds2021prompt}. The model then infers from these examples to perform the task. Since examples are presented in natural language, they provide an accessible way to engage with language models and facilitate the incorporation of human knowledge into these models through demonstrations and templates \parencite{brown2020language, dong2022survey, liu2023pre}. Currently, there is no universally standardized methodology for providing examples in shot-prompting \parencite{Anil_2023, Dragon_2023, Tam_2023}. For more straightforward tasks, like language translation or classification, a prompt could be formulated as demonstrated in Listing \ref{lst:promptshoteasy}. For more complex tasks, like content generation, a prompt can be constructed as demonstrated in Listing \ref{lst:promptshotcomplex}.


\begin{lstlisting}[caption=Example of few-shot prompt in a straightforward task., label=lst:promptshoteasy]
Text: My ear feels fine after the treatment.
Classification: Positive
Text: The doctor examined my ear, and everything seems normal.
Classification: Neutral
Text: I experience some discomfort, I suspect it might be ear infection.
Classification: Negative
Text: The ear pain is unbearable,  I need to see a specialist.
Classification:
\end{lstlisting}

\begin{lstlisting}[caption=Example of few-shot prompt in a complex task., label=lst:promptshotcomplex]
Write a medical report about the following transcript: [transcript].
Use the following SOAP reports [example report 1] and [example report 2] as a guide.
\end{lstlisting}

\subsubsection{Pattern Prompting}
\textit{Pattern prompting} involves the availability of various patterns that can be chosen and employed as the basis for the formulation of prompts. These patterns facilitate interactions with conversational LLMs across various contexts, extending beyond just discussing interesting examples or domain-specific prompts \parencite{white2023prompt}. The aim is to codify this knowledge into pattern structures that enhance the ability to apply it in different contexts and domains where users encounter similar challenges, although not necessarily identical ones. This approach promotes greater reuse and adaptability of these patterns for diverse use cases and situations \parencite{white2023prompt}.

The study of \cite{white2023prompt} introduces, among others, the \textit{context control pattern} category. Context control captures the \textit{context manager pattern}, which enables users to specify or remove context from the prompt. ``By focusing on explicit contextual statements or removing irrelevant statements, users can help the LLM better understand the question and generate more accurate responses'' \parencite{white2023prompt}. The greater the clarity in the statements, the higher the likelihood that the LLM will respond with the intended action. Possible context statements are: ``within the scope of X'', ``consider Y'', ``ignore Z''; an example is shown in Listing \ref{lst:promptcontext} \parencite{white2023prompt}. 

\begin{lstlisting}[caption=Example of a prompt using the context manager pattern., label=lst:promptcontext]
Listen to this transcript between doctor and patient and make a EHR entry from it.
Consider the medical guidelines.
Do not consider irrelevant statements.
\end{lstlisting}

\subsection{Automatic Text Summarization}
Since the introduction of transformer-based methods in ATS, the usage of prompt engineering has been instrumental in enhancing the performance of ATS processes. In ATS, various pragmatic algorithms can be integrated into computers to generate concise summaries of information \parencite{mridha2021survey}. When used in Natural Language Processing (NLP), ATS is used to evaluate, comprehend, and extract information from human language \parencite{mridha2021survey}. The introduction of transformer-based models like GPT \parencite{radford2019language} shows improved performance in NLP-tasks \parencite{mridha2021survey} which is beneficial for abstractive summarization.  
 
Abstractive summarization creates summaries by introducing new phrases or words not present in the original text. To achieve accurate abstractive summaries, the model must thoroughly comprehend the document and express that comprehension concisely through new terms or alternative expressions \parencite{widyassari2019literature}. The opposite of abstractive summarization, is extractive summarization, a method where the summary consists entirely of extracted content \parencite{widyassari2019literature}. Extractive summarization has been used most frequently because it is easier, but the summaries generated are far from human-made summaries, in contrast to abstractive summarization \parencite{widyassari2019literature,yadav2022automatic}.

\subsection{Medical Dialogue Summarization}
In MDS, it is important that the summaries are at least partly abstractive. In one respect, the reports are generated from dialogue, so extracting literal (sub-)sentences will not lead to a coherent report; conversely, the summaries must be comparable to the human-made versions of the general practitioners (GP). In MDS, the relevant medical facts, information, symptoms, and diagnosis must be retrieved from the dialogue and presented either in the form of structured notes or unstructured summaries \parencite{jain2022survey}. The most common type of medical notes are SOAP notes: \textbf{S}ubjective information reported by the patient, \textbf{O}bjective observations, \textbf{A}ssessment by medical professional and future \textbf{P}lans \parencite{jain2022survey,krishna2020generating}. \\

\begin{table*}[t]
    \small
    \centering
    \caption{Example of part of a consultation transcript and the corresponding SOAP report (translated to English, the original transcript and SOAP report are in Dutch).}
    \label{tab:report}
    
    \begin{tabular}{ p{0.5\textwidth}| p{0.45\textwidth}}
    \textbf{Transcript} & \textbf{SOAP report} \\
    \hline
    \textbf{GP:} Good morning. & \multirow{6}{0.98\columnwidth}{\textbf{S:} Since 1.5 weeks, ear pain and a feeling of deafness right ear, received antibiotics from the GP. Feeling sicker since yesterday, experiencing many side effects from the antibiotics. Using Rhinocort daily for hyperreactivity. Left ear operated for cholesteatoma, no complaints.} \\

    \textbf{P: }Good morning, hello. Last week I visited your colleague. \\
    \textbf{GP:} Yes I see, for your ear. & \\
    \textbf{P:} I had an ear infection. Well, I'm actually getting sicker. Since yesterday, I've been getting sicker and sicker. & \\
    \textbf{GP:} She gave you antibiotics, right? & \\
    \textbf{P:} Yes, the first three or four tablets were really like, whoa. And after that, it was just the same. So, I still have ear pain. And now I notice that my resistance is decreasing because of &  \multirow{2}{0.97\columnwidth}{\textbf{O:} right ear: eardrum visible, air bubbles visible, no signs of infection. [left ear ?]} \\
    the antibiotics. I'm just getting more tired now. & \textbf{A:}  OMA right \\
    \tiny ... &   \\
    \textbf{GP:} We're just going to take a look. There is some fluid. Also, air bubbles behind the eardrum. That is clearly visible. & \multirow{2}{\columnwidth}{\textbf{P:} Advice xylomethazine 1 wk, continue antibiotics, review symptoms in 1 week. Consider prescribing Flixonase, referral to ENT?} \\
    \textbf{P:} Yes, yes, yes, that's correct. It gurgles and it rattles and it & \\
     rings. And it's just blocked. &  \\
    \textbf{GP:} Yes, I believe that when I see it like this. It doesn't look & \\
    red. It doesn't appear to be really inflamed. & \\
    \tiny ... & \\
    \textbf{GP: }I think, for now, at least, you should finish the antibiotics. & \\
    \textbf{P:} That's two more days. &  \\
    \textbf{GP:} Yes, and continue using the nasal spray, or the other nasal spray, for another week and see how it goes. Just come back if it's still not better after a week. And if it persists, well, maybe then you should see the ENT specialist. & \\
    \tiny ... & \\
    
    \multicolumn{1}{l}{GP = General Practitioner, P = Patient} & \multicolumn{1}{l}{OMA = Otitis Media Externa, ENT = Ear, Nose, Throat} \\

    \end{tabular}

\end{table*}


Previous work in MDS has produced the transformer-based approaches of MEDSUM-ENT \parencite{nair2023generating}, MedicalSum \parencite{michalopoulos2022medicalsum}, and SummQA \parencite{mathur2023summqa}. 

\newpage

\begin{itemize}
    \item ``MEDSUM-ENT is a medical conversation summarization model that takes a multi-stage approach to summarization, using GPT-3 as the backbone''. MEDSUM-ENT first extracts medical entities and their affirmations and then includes these extractions as additional input that informs the final summarization step through prompt chaining. Additionally, MEDSUM-ENT exploits few-shot prompting for medical concept extraction and summarization through in-context example selection. Their study concludes that summaries generated using this approach are clinically accurate and preferable to naive zero-shot summarization with GPT-3 \parencite{nair2023generating}. 
    \item MedicalSum is a sequence-to-sequence architecture for summarizing medical conversations by integrating medical domain knowledge from the Unified Medical Language System (UMLS) to increase the likelihood of relevant medical facts being included in the summarized output. Their analysis shows that MedicalSum produces accurate AI-generated medical documentation \parencite{michalopoulos2022medicalsum}.
    \item SummQA is a ``two-stage process of selecting semantically similar dialogues and using the top-k similar dialogues as in-context examples for GPT-4''. They generate section-wise summaries and classify these summaries into appropriate section headers. Their results highlight the effectiveness of few-shot prompting for this task \parencite{mathur2023summqa}.
\end{itemize}

The present study not only builds upon this existing knowledge base by integrating a combination of shot prompting and context patterns into prompt engineering but also includes a crucial human evaluation component, in addition to the accuracy measurement. This human evaluation provides comprehensive insights into prompt performance beyond computer-based metrics. Leveraging GPT-4 for Dutch consultations, we ensure that the resulting medical reports adhere to the widely recognized SOAP guidelines. It is noteworthy that while prior studies have demonstrated the efficacy of shot-prompting, this is the first published study to incorporate both shot prompting and context pattern prompting in the domain of Dutch MDS, thereby making a significant contribution to the Dutch medical field.

\section{\uppercase{Study Design}}
\label{sec:studydesign}
\begin{figure}[b]
\centerline{
\includegraphics[width=\linewidth]{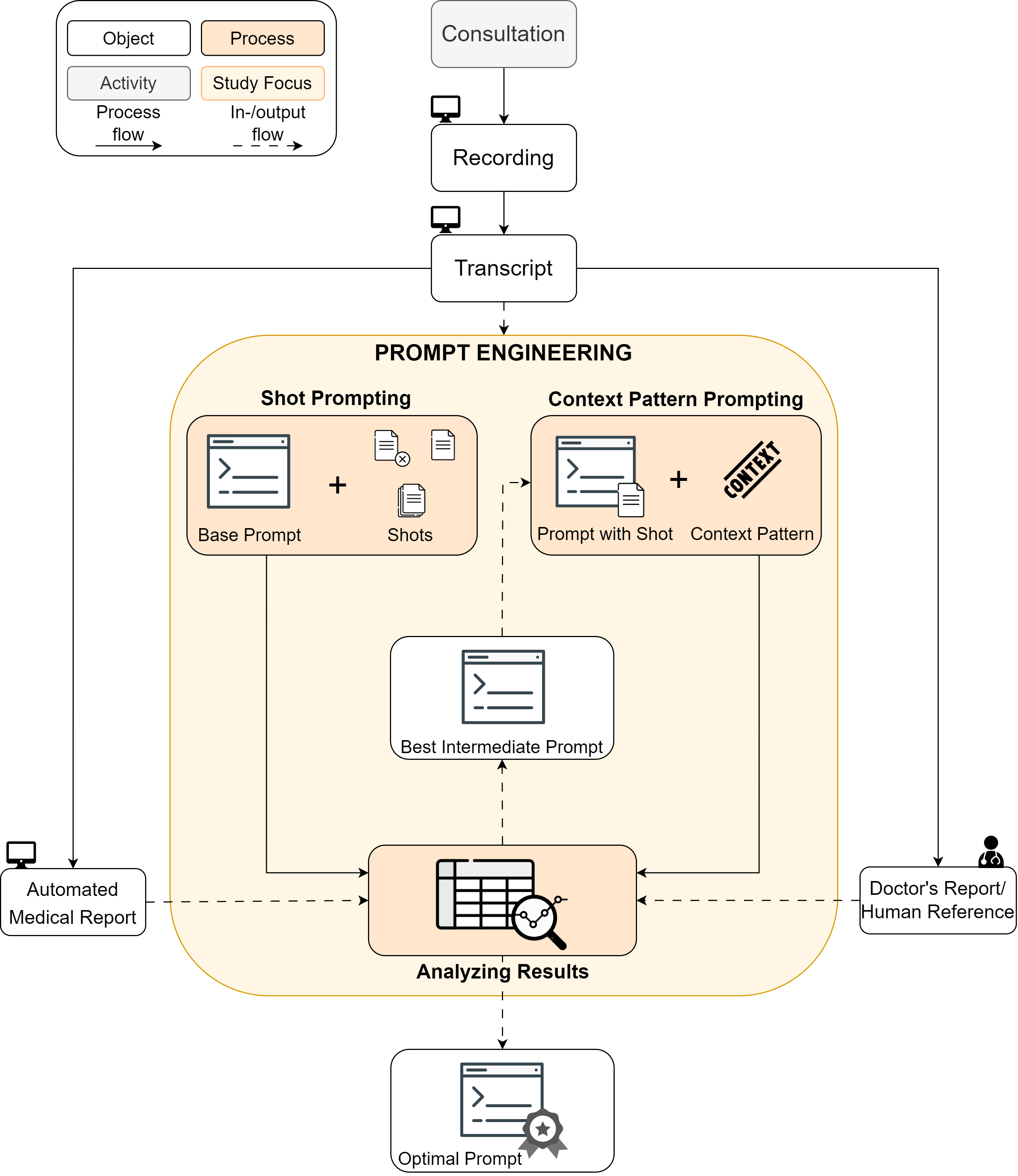}}
\caption{Research method visualization.}
\label{fig:RCD}
\end{figure}

We conducted a causal-comparative study to identify the cause-effect relationship between the formulation detail of the prompt and the performance of the automated medical report \parencite{schenker2004causal}. We followed the approach of the C2R program, by using transcripts that were made of the verbal interaction during a series of video-recorded consultations between GPs and their patients (Figure \ref{fig:RCD}) \parencite{maas2020care2report, meijers2019shared}. The recordings, for which patients as well as GPs provided informed consent, were made as part of previous communication projects carried out by researchers at Radboudumc and Nivel (Netherlands institute for health services research) \parencite{houwen2017improving}. Subsequently, medical professionals examined these transcripts to generate SOAP medical reports, with an illustrative example presented in Table \ref{tab:report}. These SOAP reports are used in the study as a human reference for comparison with the automatically generated reports. The automatically generated medical reports were produced by GPT based on various prompt formulations. Using prompt engineering, the prompts were created using the \textit{shot prompting} and \textit{context manager pattern} techniques. Each executed prompt resulted in medical reports that were analyzed to determine which prompt yielded the best results.

\subsection{Formulation of Prompts}
\begin{figure}[t]
\centerline{
\centering
\includegraphics[width=1.0\linewidth]{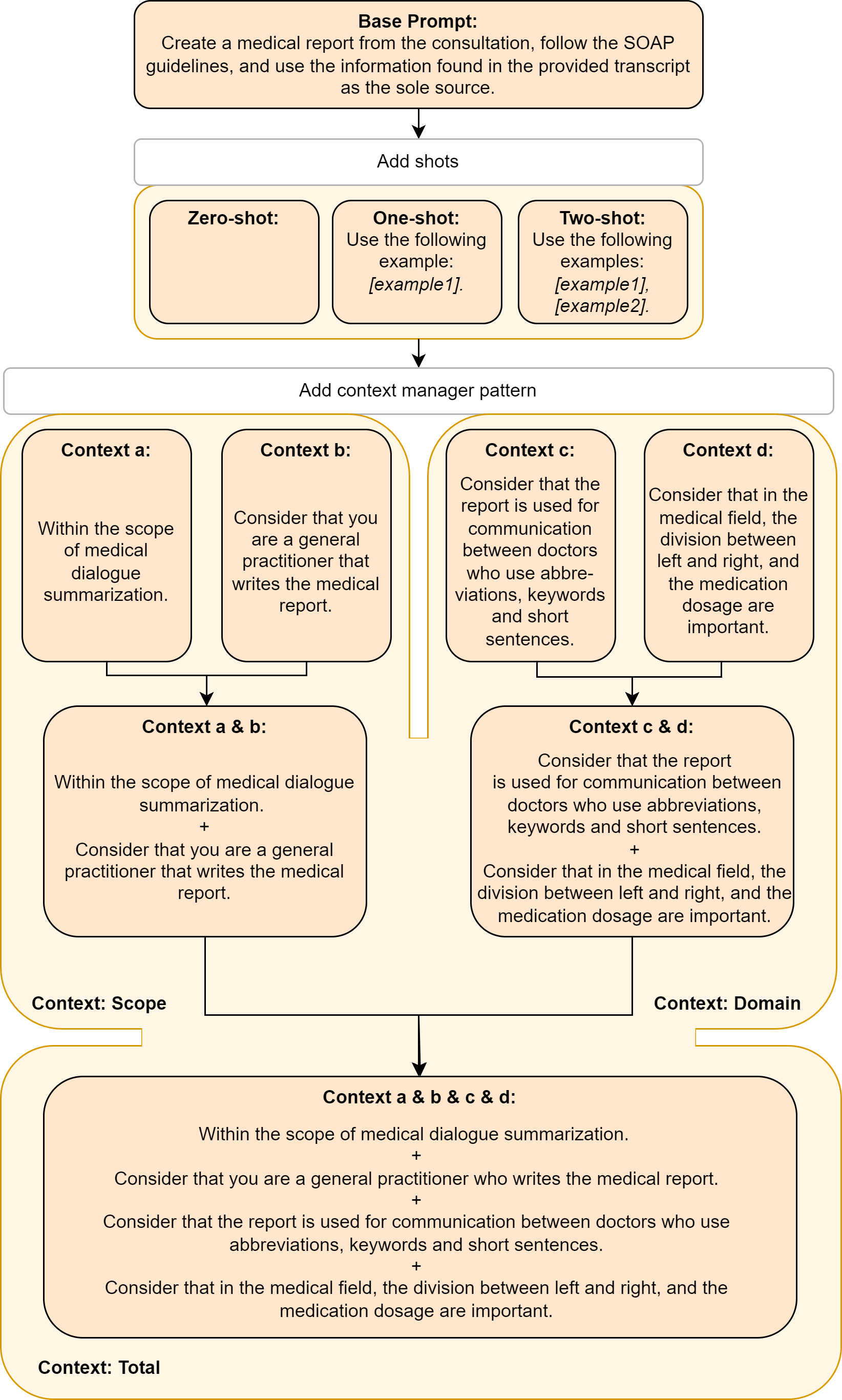}}
\caption{The flow of prompt formulation (translated to English, the original prompts are in Dutch).}
\label{fig:PF}
\end{figure}

The prompts formulated in this work combine \textit{shot prompting} and \textit{context pattern prompting} (Figure \ref{fig:PF}). First, a base prompt was established upon which all other elements in the prompt could be built. Variability in performance can then be attributed solely to differences in shots or context, rather than possible other factors. The base prompt compels the GPT to solely utilize elements present in the transcript to prevent hallucinations \parencite{banerjee2023benchmarking,ji2023survey}.

This base prompt was initially employed to construct three versions of \textit{shot-prompting}: zero-shot, one-shot, and two-shot. The most effective shot-prompting among these three prompts was selected. Using the \textit{context manager pattern}, an increase in context was added to the prompt to measure the effect of incorporating more context into the prompt. The context is divided into two types of contexts: scope context and domain context. 
The scope context explains in what scope the GPT operates and what its role is. The domain context gives more details about communication and important elements in the medical field.


The following context statements are included:

\begin{itemize}
    \item[a.] Within the scope of medical dialogue summarization;
    \item[b.] Consider that you are a general practitioner who writes the medical report during the consultation;
    \item[c.] Consider that the report is used for communication between doctors who use abbreviations and short sentences or keywords;
    \item[d.] Consider that in the medical field, the division between left and right, and the medication dosage are important.
\end{itemize}

Each of these statements, as well as various combinations of them, as illustrated in Figure \ref{fig:PF}, were included to assess their individual, as well as their combined effects.

\subsection{Running of Prompts}
The crafted prompts served as a means to collect and assess the data concerning their performance in practice. The formulated prompts were run in a self-written prompt engineering software supported by the Azure OpenAI Service, which is a fully managed service that allows developers to easily integrate OpenAI models into applications \parencite{Mickey_2023}. GPT-4 was used with a temperature of 0; GPT-4 is the current best-performing GPT-version and a temperature of 0 minimizes the creativity and diversity of the text generated by the GPT model \parencite{liu2023gpteval}. 

As a data source, seven real-world Dutch consultations between a GP and patients, concerning Otitis Externa and Otitis Media Acuta, were utilized and employed in three distinct manners:
\begin{itemize}
    \item Five transcriptions of these consultations served as input data to create automated medical reports. 
    \item Five manually created SOAP reports (of these five transcriptions) by doctors were employed as a human reference for the automated medical reports.
    \item Two manually created SOAP reports by doctors were used as examples in shot-prompting. 
\end{itemize}

It has been ensured that both an external and middle ear infection consultation are included in the examples, but the distinction between input and example data has been randomly made. On average, the dialogue transcriptions consisted of $1209$ words ($SD = 411$), ranging between $606$ and $1869$ words. The manually created SOAP reports consisted, on average, of $60$ words ($SD = 17$), ranging between $37$ and $87$ words. 

\subsection{Analysis of Prompts}
Despite growing interest in ML as a tool for medicine, there is a lack of knowledge about how these models operate and how to properly assess them using various metrics \parencite{hicks2022evaluation}. In this study, the resulting automated reports were evaluated against the human reference reports using accuracy metrics and a human evaluation. By combining these quantitative and qualitative insights, this two-step review approach gives a comprehensive assessment of the automated reports' performance. 

\subsubsection{Accuracy Measurement}
We used ROUGE as an accuracy metric since this is the most used text summarization evaluation metric to automatically evaluate the quality of a generated summary by comparing it to a human reference and it is suitable for our Dutch reports \parencite{barbella2022rouge, tangsali2022abstractive}. The ROUGE metric code offered by the HuggingFace library was used to calculate the ROUGE1, and ROUGEL scores of the automated medical reports \parencite{lin-2004-rouge}. ROUGE1 assessed the unigram similarities, and ROUGEL the longest common subsequence of words, between the automated report and the human reference reports  \parencite{tangsali2022abstractive}. 

The generation of the automated medical report is stochastic because generative AI models frequently display variety in their replies to a given prompt. To account for this variability every prompt was run five times on all transcripts, yielding distinct responses with each run. The prompts were run five times to strike a balance between robustness and computational efficiency, taking the trade-off between thorough analysis and computational costs into account. For every run, ROUGE was calculated. The overall performance and consistency of the automated medical reports are indicated by computing the average ROUGE score per consultation. Finally, an overall mean of these averages, with their standard deviations, was calculated and presented in the findings. 

\subsubsection{Human Evaluation}
It is important to note that none of the automatic evaluation metrics are perfect and that human evaluation is still essential to ensure the quality of generated summaries \parencite{Falcão_2023}. For the human evaluation, the generated reports were manually analyzed. The words in the reports were categorized into three groups based on whether they were identical, paraphrased, or additional to the human reference. We also identified and classified the additional statements in the automatic reports into distinct categories. The identified categories were: duration of complaints, duration of treatment, previously tried treatments, doctor's observations, specific complaints (all reported symptoms by the patient), referral to which hospital, wait for results, discussed treatment (all specific steps that the GP reports to the patient), expected patient actions, and other complaints that are ultimately not related to the diagnosis made in the human reference. Based on the clinical report idea of \cite{savkov2022consultation} six medical professionals were asked to evaluate the importance of these classified additions in a SOAP report. Based on the response of the medical professionals, the additions were classified according to an adapted version of the taxonomy of error types by \cite{moramarco2022human}. Not all of their errors were observed in our study, besides, we replaced their ``incorrect order of statements error'' with a ``categorization error'', and we identified ``redundant'' statements additionally. 

\section{\uppercase{Findings and Discussion}}
\label{sec:findingsanddiscussion}
Running of the formulated prompts, resulted in automated medical reports with wordcounts shown in Table \ref{tab:wordcount}. The automated reports are approximately twice as long as the human references, indicating a significant disparity in the length of the generated content. This could be explained by the fact that GPT generates full sentences, providing more detailed descriptions, while GPs tend to use abbreviations and keywords to convey the same information more concisely. Four out of six GPs in the expert panel indicated that they prefer abbreviations and keywords over full sentences, however, one GP preferred full sentences. 

\begin{table}[b]
    \centering
    \small
    \caption{Word count comparison between the generated report and the human reference.}
    \label{tab:wordcount}
    \begin{tabular}{l c c c}
    \hline
    & \multicolumn{1}{c}{\textbf{Human}} & \multicolumn{1}{c}{\textbf{Generated}} & \textbf{Difference} \\
    & \multicolumn{1}{c}{\textbf{Reference}} & \multicolumn{1}{c}{\textbf{Report}} \\
    \hline
      \textbf{S}ubjective & $29$ & $47$ & $18$ \\
      \textbf{O}bjective & $11$ & $20$ & $9$ \\
      \textbf{A}nalysis & $4$ & $8$ & $4$ \\
      \textbf{P}lan & $14$ & $33$ & $19$ \\
    \hline
      \textbf{Total} & $111$ & $58$ & $53$ \\ 
    \hline
    \end{tabular}

\end{table} 

\subsection{Accuracy Measurement}
In the evaluation of the accuracy of the prompts, first, the \textit{shot-prompting} technique was evaluated, followed by the \textit{context manager pattern} technique that built on the optimal numbers of shots.

\subsubsection{Shot-prompting}
Table \ref{tab:shot-results} shows the comparison of the different shot-prompting approaches. The comparison shows that the \textit{zero-shot prompting} approach resulted in the lowest ROUGE scores ($0.121$ and $0.079$). \textit{One-shot prompting} resulted in slightly higher scores ($0.150$ and $0.104$) and the \textit{two-shot prompting} approach resulted in the highest ROUGE scores ($0.174$ and $0.123$). This result shows that adding shots to a prompt improves the performance. This can be explained by the fact that the shots serve as a reference for the expected output, enabling the GPT to generate similar outputs, which is in line with earlier research \parencite{reynolds2021prompt}. 
Adding an increasing number of shots could result in higher performances than two-shots since few-shot prompting is generally meant to include a larger set of examples \parencite{brown2020language}. This was, however, not possible due to the limited data set. Controversially, using fewer examples, makes it possible to create more well-crafted examples and comes closer to human performance \parencite{brown2020language}. Additionally, \cite{zhao2021calibrate} found that few-shot prompting might introduce biases into certain answers.

\begin{table}[t]
\small
\centering
\caption{Mean and Standard Deviation (SD) for the ROUGE1 and ROUGEL-scores for zero-shot, one-shot, and two-shot prompting.}
\label{tab:shot-results}
\begin{tabular}{p{0.34\linewidth} c c } 
\hline
& \textbf{ROUGE1}& \textbf{ROUGEL} \\ 
& Mean±SD & Mean±SD \\
\hline
\textbf{Zero-shot} & $0.121$\textcolor{gray}{±$0.007$} & $0.079$\textcolor{gray}{±$0.006$} \\
\textbf{One-shot} & $0.150$\textcolor{gray}{±$0.009$} & $0.104$\textcolor{gray}{±$0.006$} \\
\textbf{Two-shot} & $0.174$\textcolor{gray}{±$0.005$} & $0.123$\textcolor{gray}{±$0.004$}  \\
\hline
\end{tabular}
\end{table}

It is also worth considering that the absence of a universally accepted method for applying shot prompting introduces a degree of uncertainty regarding the most effective approach. Including the transcripts with the sample SOAP reports, rather than only presenting the SOAP report as an example could have potentially produced different results. However, it is important to note that the main goal of this study was to teach the GPT how to correctly use the SOAP format and how to describe items in the SOAP categories.     

\subsubsection{Context manager pattern}
The two-shot prompting strategy produced the highest scores, thus the \textit{context manager pattern} was added to this foundation. Scope context and domain context were evaluated separately as well as the combination of the two types of context. In the assessment of the context manager pattern, a slight variation could be observed in the ROUGE scores based on different contextual additions (Table \ref{tab:context-results}). 

\begin{table}[t]
\small
\centering
\caption{Mean and Standard Deviation (SD) for the ROUGE1 and ROUGEL-scores for context prompts.}
\label{tab:context-results}
\begin{tabular}{l c c}
\hline
& \textbf{ROUGE1} & \textbf{ROUGEL} \\ 
& Mean±SD & Mean±SD \\
\hline
\multicolumn{3}{l}{\textbf{Context: Scope}} \\
\hline
\textbf{Context a} & $0.172$\textcolor{gray}{±$0.041$} & $0.120$\textcolor{gray}{±$0.016$}  \\
\textbf{Context b} & $0.173$\textcolor{gray}{±$0.043$} & $0.124$\textcolor{gray}{±$0.022$}  \\
\textbf{Context a \& b} & $0.179$\textcolor{gray}{±$0.049$} & $0.126$\textcolor{gray}{±$0.023$}  \\
\hline
\multicolumn{3}{l}{\textbf{Context: Domain}} \\
\hline
\textbf{Context c} & $0.242$\textcolor{gray}{±$0.035$} & $0.179$\textcolor{gray}{±$0.016$}  \\
\textbf{Context d} & $0.173$\textcolor{gray}{±$0.048$} & $0.121$\textcolor{gray}{±$0.025$}  \\
\textbf{Context c \& d} & $0.220$\textcolor{gray}{±$0.064$} & $0.167$\textcolor{gray}{±$0.037$}  \\
\hline
\multicolumn{3}{l}{\textbf{Context: Total}} \\
\hline
\textbf{Context a \& b \& c \& d} & $0.250$\textcolor{gray}{±$0.049$} & $0.189$\textcolor{gray}{±$0.025$}  \\
\hline
\end{tabular}
\end{table}

The combined \textit{scope context} ($0.179$ and $0.126$) scored lower than the combined \textit{domain context} ($0.220$ and $0.167$). This would suggest that \textit{scope context} has little effect on the quality of reports that are generated. However, the combination of \textit{scope context} and \textit{domain context} ($0.250$ and $0.189$) resulted in higher ROUGE scores than \textit{domain context} by itself. A noteworthy finding is the difference between \textit{domain contexts c} and \textit{domain context d}, where \textit{context d} produced lower scores ($0.173$ and $0.121$) than \textit{context c} ($0.242$ and $0.179$). Remarkably, \textit{contexts c} and \textit{context d} together ($0.220$ and $0.167$) also produced lower results than \textit{context c} by itself. This suggests a potential negative effect of \textit{context d} on the overall performance. To test this, a prompt was run that excluded \textit{context d} from the prompt but this led to even lower overall scores ($0.239$ and $0.178$). This decline in score may be explained by the limited dataset, which could have resulted in skewed results. 

A potential reason why domain context increases the performance more than scope context is that the shot prompting already provides clear direction on how the GPT should behave; it has already set the context to the medical field. Prompting to use abbreviations, short sentences, and keywords (\textit{context c}), may have had a considerable influence since GPT itself tends to make long sentences and provide as much information as possible. Prohibiting this action resulted in improved performance in the automated report. It is notable that it is unexpected that the GPT does not already do this after the shot prompting, but this could possibly be explained because only SOAP examples were used without including the transcripts in the examples.     

This study also investigated the inclusion of a list of abbreviations within the prompt and found that it had a positive impact on the results, with ROUGE scores of $0.273$ and $0.261$. However, it was ultimately not selected as the optimal prompt since the use of abbreviations varies between hospitals and healthcare providers \parencite{borcherding2007ota}, making it difficult to create a universally applicable prompt that incorporates all relevant abbreviations.

\begin{table*}[t]
    \small
    \centering
    
    \caption{Error statements with occurrences in the five generated medical reports (translated to English, the generated reports are in Dutch).}
    \label{tab:humanevaluation}
    
    \begin{tabular}{l p{0.60\textwidth} c} 
    \hline
    \textbf{Type} & \textbf{Definition - Examples} & \textbf{Occurrence} \\
    \hline
    \textbf{Factual Errors} & An error in the information presented that contradicts reality. & $14$ \\
    \hline
    Hallucinations & \textit{``Pain originating from the syringing by the doctor's assistant''} & $6$ \\
    & Pain was already present before the syringing. &  \\
    Incorrect statements & \textit{``The patient uses Rhinocort and cetirizine daily for mucous membrane hyperreactivity''} & $8$ \\
    & Patient only uses Rhinocort for mucous membrane hyperactivity. &  \\
    \hline
    \textbf{Stylistic Errors} & An error in the manner in which information is used or presented. & $17$\\
    \hline
    Repetitions & \textit{``Patient feels sick''} & $3$ \\
    & \textit{``Patient also reports a feeling of being unwell''}. &  \\
    Classification error & \textit{``The area around the ear feels numb.''} & $14$ \\
    & in the \textbf{A}nalysis part of SOAP. & \\
    \hline
    \textbf{Omissions} & An error characterized by the act of neglecting to include essential information in the report. & $19$ \\
    \hline
    In \textbf{S}ubjective &  Indication of which ear is involved/ referred to & $3$ \\
    & Parts of symptoms mentioned & $2$ \\
    & Parts of relevant medical history & $5$ \\
    In \textbf{O}bjective & Indication of which ear is involved/ referred to & $2$ \\
    & Parts of symptoms observed & $2$ \\
    In \textbf{A}nalysis & Indication of which ear is involved/ referred to & $3$ \\
    In \textbf{P}lan & Agreement with patient & $1$ \\
    & Possible future treatment & $1$ \\
    
    \hline
    \textbf{Redundant Statements} & The inclusion of unnecessary information that does not contribute substantively to the report, although it is on the topic of the medical condition. & $25$ \\
    \hline
    In \textbf{S}ubjective & \textit{The patient reports ... especially in the morning, and that the ear smells.} & $7$ \\
    In \textbf{O}bjective & \textit{Left: some earwax.} & $5$ \\
    In \textbf{A}nalysis & \textit{This can also radiate from the sinuses.} & $2$ \\ 
    In \textbf{P}lan & \textit{A dressing and plaster have been applied to the left ear to collect the discharge.} & $9$ \\
    Additional & \textit{Colonoscopy scheduled for three years. Patient should contact for referral to a gastroenterologist. Prescription for [name of medication] for constipation.} & $2$ \\
    & In an additional \textit{NB} (Nota Bene) & \\
    \hline
    \multicolumn{3}{p{0.95\textwidth}}{The occurrence is counted per consultation, so if the same error happened repeatedly in the reruns for the same consultation, it was only counted once.}
    \end{tabular}
\end{table*}

\subsection{Human Evaluation}
The results from the quantitative approach showed that the \textit{two-shot prompting} approach in combination with the \textit{scope} and \textit{domain context} (Listing \ref{lst:finalprompt}) resulted in the best performance. However, since this still resulted in a relatively low ROUGE score, human evaluation was performed for this final prompt.

\begin{lstlisting}[caption=The best performing prompt., label=lst:finalprompt]
Within the scope of medical dialogue summarization, create a medical report from the consultation, follow the SOAP guidelines, and use the information found in the provided transcript as the sole source. 
Consider that you are a general practitioner who writes the medical report.
Consider that the report is used for communication between doctors who use abbreviations and short sentences.
Consider that in the medical field, the division between left and right, and the medication dosage are important.
Use the following examples: [example1], [example2].
\end{lstlisting}

\begin{figure}[b]
\centerline{
\includegraphics[width=0.95\linewidth]{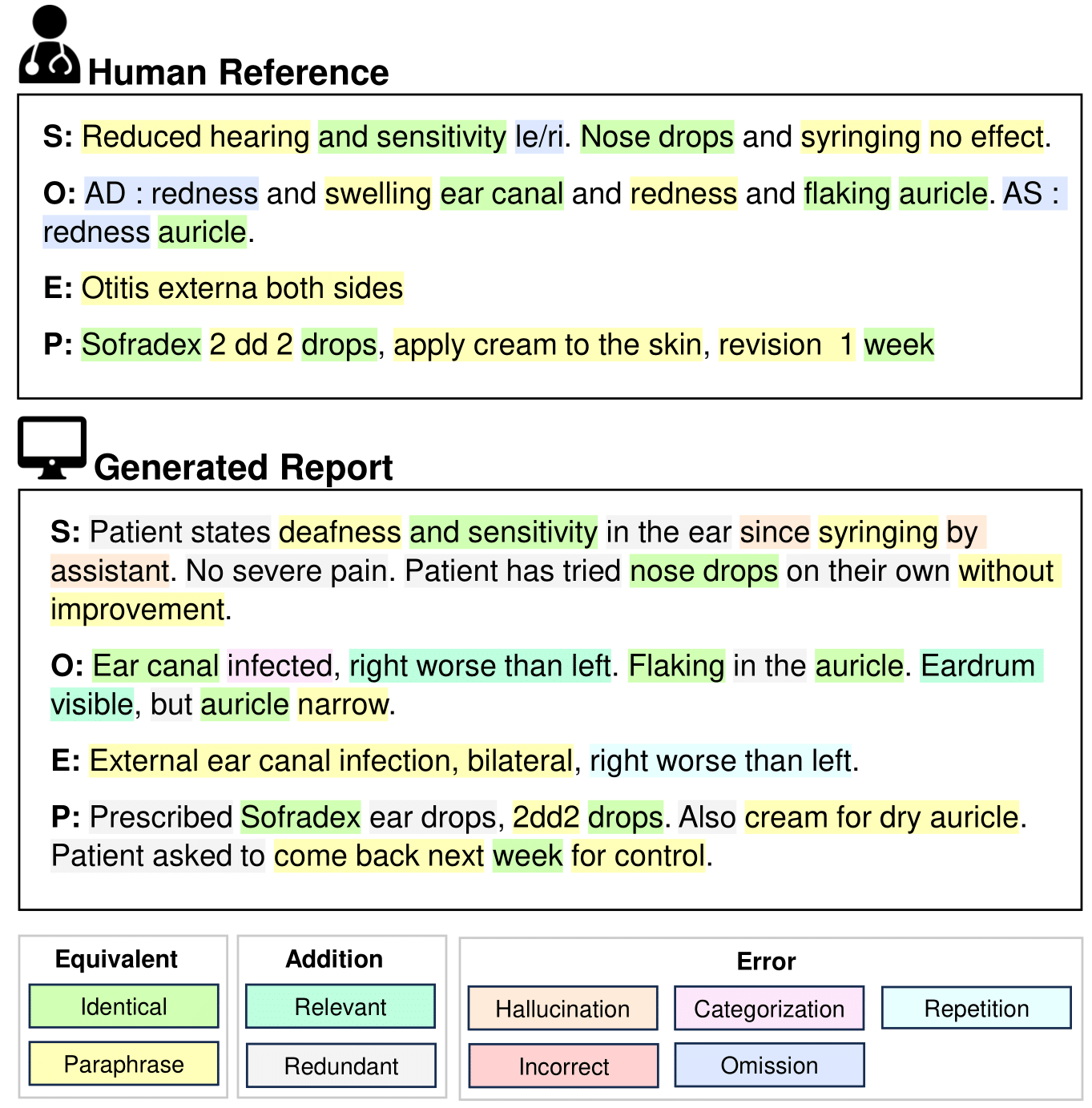}}
\caption{Human evaluation of the automated medical report of transcript 2028 (translated to English, the generated reports are in Dutch).}
\label{fig:HE2}
\end{figure}

The expert panel showed that all six GPs agreed on the fact that the duration of the complaints is relevant to mention within the report. For all the other categories there seems to be disagreement about the relevance. For example, there appears to be disagreement about the importance of recording specific patient complaints. When mentioning that in particular, the left ear caused problems, the GPs disagree on the importance. Some indicate that this is relevant ($n = 3$), while there are also GPs that indicate that this is not relevant ($n = 1$) or they indicate that they are neutral about this ($n = 2$). However, one of the GPs who indicated that it is relevant did mention that they would note it more briefly. Another example that shows this disagreement is within the discussed treatment: ``gauze and plaster applied to the left ear to collect discharge''. Two GPs indicated that this was relevant, two indicated that this was irrelevant and two indicated that they were neutral about this.

\begin{figure}[t]
\centerline{
\includegraphics[width=0.95\linewidth]{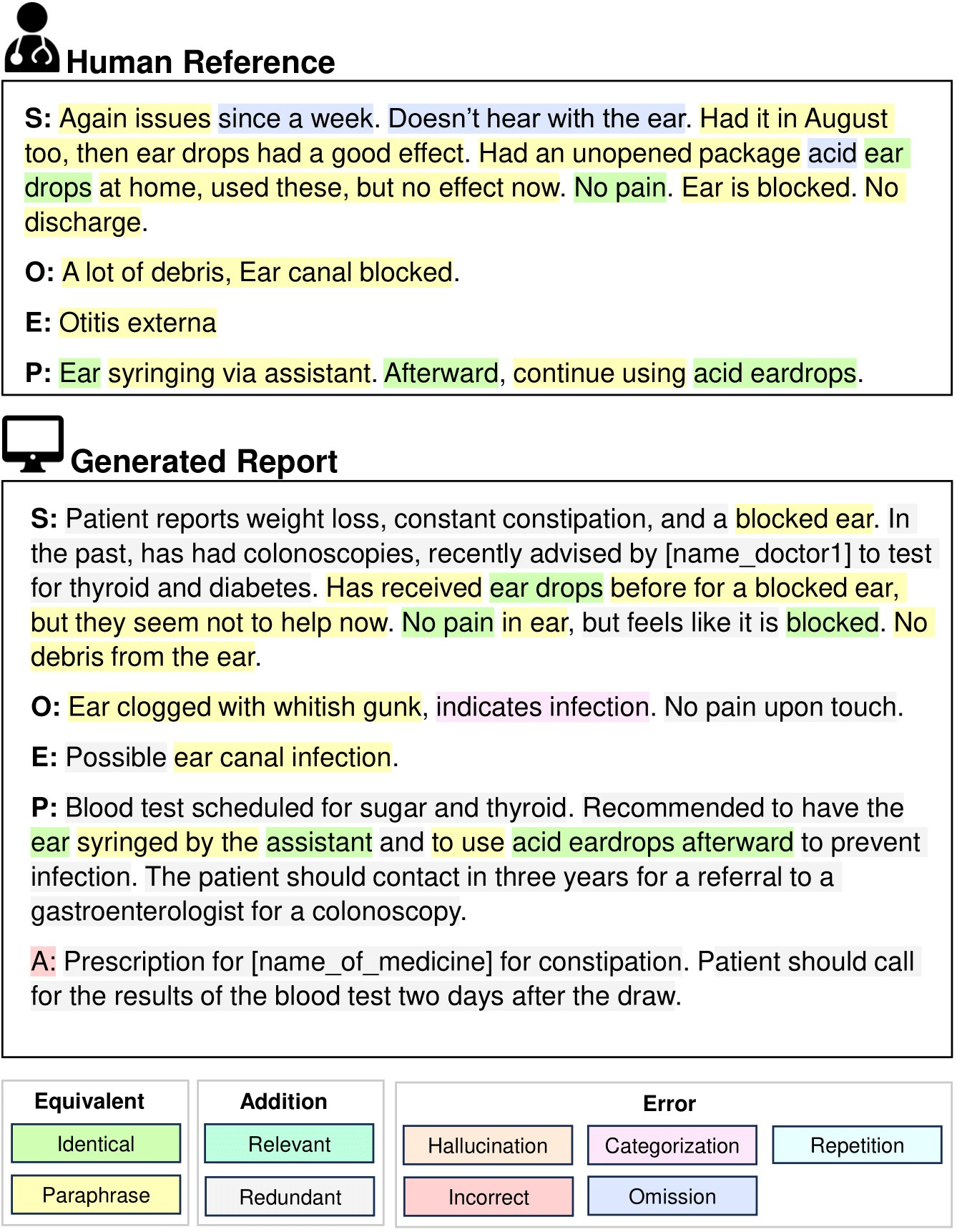}}
\caption{Human evaluation of the automated medical report of transcript 2006 (translated to English, the generated reports are in Dutch).}
\label{fig:HE10}
\end{figure}

Table \ref{tab:humanevaluation} shows the identified error statements in the five automated reports during the human evaluation. The human evaluation highlights several noteworthy findings regarding the quality of the automated reports. It is evident that the automated reports contain a notable number of redundant statements, $25$ in total, with the majority occurring in the \textbf{P}lan section ($n = 9$) and the \textbf{S}ubjective section ($n = 7$). Moreover, stylistic errors are prevalent, particularly classification errors ($ n =14$) and occasional repetitions ($n = 3$). In addition to adding extra (relevant or redundant) information, the automated report sometimes omits essential information ($n = 19$) when compared to the human reference. Factual errors are present as well, amounting to a total of $8$ incorrect statements and $6$ hallucinations. For a visual example of the error statements see Figures \ref{fig:HE2} and \ref{fig:HE10}.

A possible reason for the omissions in the automated reports could be related to the GPT's limited understanding of the medical context, leading it to overlook certain critical details during the report generation process. This is supported by research from \cite{johnson2023assessing}, who found that a potential limitation of GPT is handling complex medical queries, but they did not reach statistical significance for this statement. A potential reason for the classification errors is that the GPT lacks genuine comprehension of the distinct SOAP categories, thus negatively influencing its ability to accurately allocate information to the appropriate category within the SOAP report.

The human evaluation revealed substantial variations in the performance of the automated reports across different consultations, with some reports displaying higher performance levels than others. For example, the report that was generated based on transcript 2006 (Figure \ref{fig:HE10}) had a lot of redundant information added to the report while the report based on transcript 2808 closely resembles the human references (Figure \ref{fig:HE2}). This discrepancy in performance may be related to the difficulty that GPT encounters in differentiating between various medical conditions discussed during a single consultation, which may result in the creation of SOAP reports that include data pertaining to several medical conditions.  

One noteworthy finding is that, even though the prompts make clear how important left-right orientation is, the automated reports frequently miss it. This can be explained by the findings from the research of \cite{ye2022unreliability}; in their research, they have demonstrated that adding explanations or contextual cues alone does not necessarily ensure an improved result in the final output. This underlines the problem of ensuring that complicated, contextually relevant information is consistently included in the generated reports. This disparity highlights the ongoing difficulties in optimizing automated report production for medical contexts and argues the natural language processing system's flexibility and understanding when it comes to adding important facts.

\section{\uppercase{Conclusion}}
\label{sec:conclusion}
Even though machine learning is becoming more popular as a medical tool, little is known about these models' workings or how to appropriately evaluate them using different metrics. In this research, we investigated the combination of shot-prompting with pattern prompting to enhance the performance of automated medical reporting. The automated medical reports were generated with the use of prompt engineering software. The generated reports were evaluated against human reference provided by a GP. For this evaluation, the widespread ROUGE metric was used in combination with human evaluations.
The results showed that adding examples to the prompt is beneficial for the automated medical report. It also showed that adding both scope context as well as domain context improved the performance of the automated medical report. This resulted in the overall best structure for a prompt using a base structure in combination with two shots and scope and domain context.

\subsection{Limitations}
Despite these promising results, this study has validity threats that could have influenced the findings. Firstly, generative AI systems are stochastic which introduces variability as they produce different answers each time they are run, which may impact the reliability and repeatability of the results. Secondly, the findings have limited generalizability to other medical conditions because of the constrained data availability, with a small dataset exclusively on Otitis, and the variability in medical reporting across diverse domains. An additional concern is the missed opportunity to explore every combination of shots and contexts. However, the feasibility of this approach was constrained within the scope of this study. This influenced the study's depth of analysis and its capacity to provide nuanced insights. Lastly, the human evaluation has some limitations, even though medical professionals were consulted to gather domain expertise the human evaluation was still performed by non-medical professionals. This potentially introduced a perspective misalignment that could have influenced the interpretation and assessment of the generated medical reports. 

\subsection{Future Work}
\label{sec:future work}
This marks an initial investigation into optimizing prompt sequences with a fixed LLM. Nonetheless, we acknowledge that diverse LLMs may yield different outcomes. Additionally, future studies should explore the applicability of our findings in the setting of different medical conditions and broaden the scope of the study beyond Otitis. The prompt could be further improved to avoid redundant statements by defining the maximum length of the output, using an increasing number of shots, or using a different method of shots such as providing the consultation transcript in addition to the resulting medical report. 

Furthermore, future work should focus on finding a more suitable metric to evaluate the output. In the current research, the ROUGE metric was used for the evaluation of the automated medical report as well as human evaluation. ROUGE is commonly used within summarization tasks however it has some downsides, the metric is very black and white. It does not take into account the meaning of the words in the summarization but only the occurrence of specific words. For future work a different evaluation needs to be created, this metric needs to take into account the meaning of the automated medical report, and it needs to investigate if the essence of the automated medical report matches the golden standard. This new metric needs to take into account rewording and paraphrasing so that they are not automatically considered wrong. For optimal evaluation, the complete reports should be evaluated by GPs.   

\section*{\uppercase{Acknowledgements}}
We want to thank all the GPs and other medical professionals who aided us in our human evaluation. Special thanks go to Rob Vermond for assisting with the expert panel of the GPs. Their professional insights ensured that we could execute the human evaluation. We also would like to thank Kate Labunets for providing feedback on the paper. Finally, many thanks go to Bakkenist for the support of this research project.

\AtNextBibliography{\small}

{
\printbibliography}

\end{document}